\documentclass[11pt]{article}

\usepackage{acl}

\usepackage{times}
\usepackage{latexsym}
\usepackage[T1]{fontenc}
\usepackage[utf8]{inputenc}
\usepackage{microtype}
\usepackage{inconsolata}
\usepackage{graphicx}
\usepackage{booktabs}
\usepackage{amsmath}
\usepackage{url}

\usepackage{etoolbox}

\newcommand{\BLEUenak}{48.5}
\newcommand{\CHRFenak}{55.6}

\title{TabletCraft: Bridging a 4,000-Year Cultural Gap\\with Bidirectional Akkadian NMT and Cuneiform Rendering\thanks{Accepted to the 4th Workshop on Cross-Cultural Considerations in NLP (C3NLP) at ACL 2026. Code and pretrained model: \url{https://github.com/geoffreywang1117/cuneiscribe}. Package: \texttt{pip install cuneiscribe} (\url{https://pypi.org/project/cuneiscribe/}).}}

\author{Zhaohui Wang \\
  USC Viterbi School of Engineering \\
  University of Southern California \\
  \texttt{zwang000@usc.edu}}

\begin{document}
\maketitle

\begin{abstract}
Half a million cuneiform clay tablets survive in museums worldwide, yet modern users can neither read nor write in the world's oldest writing system, leaving a 4,000-year cultural barrier that existing NLP tools have only partially addressed.
Prior work enables one-way, scholar-oriented translation from Akkadian to English, but offers no path in the reverse direction: non-specialist users cannot compose new content in cuneiform, and therefore remain passive consumers of ancient culture rather than active participants.
We present \textsc{TabletCraft}, the first open-source system that enables \textit{bidirectional} interaction with Mesopotamian writing.
Users can read ancient tablets (Akkadian$\rightarrow$English) and compose new messages as cuneiform clay tablets (English$\rightarrow$Akkadian$\rightarrow$cuneiform$\rightarrow$rendered tablet).
The system integrates a ByT5-based translation model trained on 116K bidirectional samples, a cuneiform sign converter with 14,240 mappings (95.3\% coverage), and a visual tablet renderer, packaged as a \texttt{pip}-installable toolkit with CLI and web demo.
On the held-out Akkademia validation split (2,812 samples), we report 49.1 BLEU for Ak$\rightarrow$En and 48.5 BLEU for En$\rightarrow$Ak, the first published quantitative result in the reverse direction.
\end{abstract}

\section{Introduction}

Cuneiform is humanity's earliest writing system, spanning over three millennia (c.~3400 BCE -- 75 CE) across ancient Mesopotamia \cite{walker1987cuneiform}.
An estimated 500,000 clay tablets survive, recording royal decrees, epic poetry, trade receipts, and personal letters \cite{gutherz2023translating}, forming a cultural archive of extraordinary breadth.
Yet this archive remains largely inaccessible: only a few hundred scholars worldwide can read cuneiform, and the majority of tablets remain untranslated \cite{gordin2020reading}.

The resulting cross-cultural barrier differs from those typically studied in multilingual NLP.
It is not a barrier between two living communities that might learn each other's languages, but a one-way temporal asymmetry: ancient texts can be partially decoded for modern readers, but modern users have no productive channel into the ancient script.
Existing NLP tools reinforce this asymmetry.
Recent NMT systems \cite{gutherz2023translating,sommerschield2023ml4al} translate Akkadian \textit{into} English, but no system translates English \textit{into} Akkadian or renders the result in cuneiform.
The cultural engagement they enable is therefore passive: scholars read ancient texts, but neither specialists nor the general public can compose new ones.

To our knowledge, \textsc{TabletCraft} is the first open-source system to support both directions of Akkadian translation together with cuneiform-Unicode rendering.
A student can type ``The king rules the land'' and receive a clay tablet bearing the cuneiform equivalent; a museum visitor can render their name in the world's oldest script; a researcher can back-translate English glosses into Akkadian for data augmentation.
By supporting composition in cuneiform in addition to reading, the system enables a more participatory mode of interaction with the script for non-specialist users.

\begin{figure}[t]
\centering
\includegraphics[width=0.95\columnwidth]{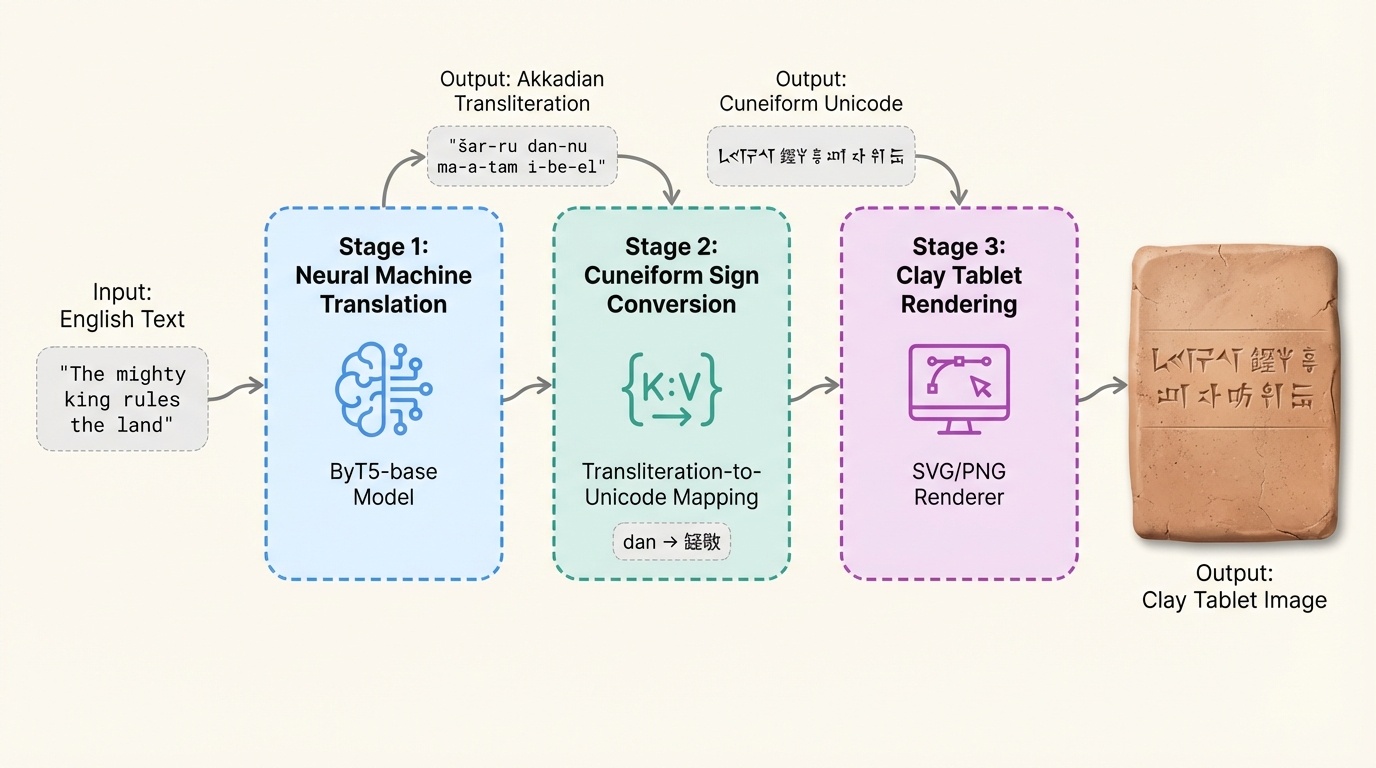}
\vspace{-6pt}
\caption{\textsc{TabletCraft} pipeline: English text is translated to Akkadian, converted to cuneiform signs, and rendered as a clay tablet, enabling bidirectional cultural interaction.}
\label{fig:pipeline}
\vspace{-8pt}
\end{figure}

\section{Cross-Cultural Design Challenges}

Ancient-language NLP raises cross-cultural problems that mainstream multilingual NLP has not yet engaged with \cite{bird2020decolonising}.
We identify four challenges that shaped \textsc{TabletCraft}'s design.

\paragraph{Temporal distance as cultural distance.}
Akkadian translation bridges a civilizational gap, not merely a linguistic one.
Concepts like ``joint-stock capital,'' ``eponym dating,'' or ``divine determinatives'' have no modern equivalents \cite{veenhof2008old}.
Our system preserves these culturally specific terms following Assyriological convention rather than forcing modernizing glosses.

\paragraph{The reverse direction as cultural agency.}
Prior systems treat ancient languages as objects of study.
The English$\rightarrow$Akkadian direction we introduce reframes users as \textit{participants}: they can compose in cuneiform, fostering a personal connection that passive reading alone cannot.
This comes with a risk of trivialization \cite{bender2021dangers}; we mitigate it by explicitly labeling all machine-generated Akkadian as approximate.

\paragraph{Accessibility vs.\ scholarly fidelity.}
We serve two distinct user populations \cite{terras2012digitisation}.
\textit{Assyriological researchers and graduate students} use the Ak$\rightarrow$En direction to triage untranslated tablets, generating draft glosses for verification rather than ground-truth translations; they expect every output to be revisable, and consume the raw transliteration string rather than the rendered tablet.
\textit{Educators, museum staff, and the general public} use the En$\rightarrow$Ak$\rightarrow$cuneiform direction for personalized engagement (a student's name on a clay tablet, a museum visitor composing a one-line dedication); they expect approximate rather than philologically exact output, and consume the tablet image with a transliteration caption.
We resolve this tension by exposing every intermediate layer (input classification, transliteration, sign mapping, rendered tablet) and by labeling all machine-generated Akkadian as approximate, rather than collapsing the two audiences into a single black-box translation.

\paragraph{Whose heritage?}
Mesopotamian heritage is disproportionately studied in Western institutions \cite{rayne2017geosciences}.
An open-source, multilingual-extensible toolkit lowers barriers for researchers and educators in Iraq, Syria, Turkey, and Iran, the modern nations whose territory encompasses ancient Mesopotamia.

\section{System Architecture}

\subsection{Translation Model}

We fine-tune ByT5-base \cite{xue2022byt5} (581M parameters) on 58,126 parallel sentences from Akkademia \cite{gutherz2023translating} (50K), a shared translation task \cite{deeppast2026} (1.5K), and sentence-aligned expansions (6K).
We train bidirectionally (Ak$\rightarrow$En and En$\rightarrow$Ak with task prefixes), yielding 116K pairs \cite{sennrich2016improving}.
Byte-level tokenization handles diacritics (\v{s}, \d{t}, \d{s}) and logograms without vocabulary mismatch \cite{lu2025byt5cuneiform}.
Training: 20 epochs, lr $10^{-4}$, batch 32, label smoothing 0.2, BF16 on 2$\times$ RTX 3090.

\subsection{Cuneiform Converter}

A lookup table of 14,240 transliteration$\rightarrow$Unicode cuneiform mappings compiled from ORACC \cite{oracc}, CDLI \cite{cdli}, and Akkademia.
Determinatives are stripped per Assyriological convention \cite{huehnergard2011grammar}.

\subsection{Tablet Renderer}

SVG/PNG output styled after Neo-Assyrian tablets (clay background, ruling lines, Noto Sans Cuneiform font). No GPU; $<$10ms per tablet.

\section{Evaluation}

\begin{table}[t]
\centering
\small
\begin{tabular}{llcc}
\toprule
\textbf{Direction} & \textbf{System} & \textbf{BLEU} & \textbf{chrF++} \\
\midrule
Ak$\rightarrow$En & Akkademia CNN$^\dagger$ & 37.5 & -- \\
Ak$\rightarrow$En & \textsc{TabletCraft} & \textbf{49.1} & \textbf{63.1} \\
\midrule
En$\rightarrow$Ak & \textsc{TabletCraft} & \BLEUenak{} & \CHRFenak{} \\
\bottomrule
\end{tabular}
\caption{Translation quality on the Akkademia validation set (2,812 Neo-Assyrian samples), reported in \emph{both} directions.
$\dagger$: \citet{gutherz2023translating}, on their own test split (different domain mix).
For the En$\rightarrow$Ak direction, references are the original Akkadian transliterations from the held-out validation split (no prior published baseline; the bidirectional model is trained to predict transliteration, not Unicode signs).}
\label{tab:results}
\end{table}

Table~\ref{tab:results} reports bidirectional translation quality.
Both directions are evaluated on the same 2{,}812-sample Akkademia validation split, which is held out from training and consists primarily of Neo-Assyrian royal inscriptions.
Direct comparison with prior work on different test splits or text genres (e.g., Old Assyrian trade correspondence) requires caution \cite{veenhof2008old}.

\paragraph{En$\rightarrow$Ak evaluation methodology.}
We provide the first quantitative measure of English$\rightarrow$Akkadian translation quality at this scale.
The bidirectional model is trained with task prefixes (\texttt{translate Akkadian to English:} and \texttt{translate English to Akkadian:}) on 116K pairs derived from 58K aligned sentences \cite{sennrich2016improving}.
We invert the validation split, using \texttt{valid.en} as source and \texttt{valid.tr} (the original transliteration) as reference, apply the same diacritic-and-gap normalization used during training, and decode with 4-way beam search.
Scores are computed with sacreBLEU.\footnote{\texttt{nrefs:1|case:mixed|smooth:exp|tok:13a}; chrF++: \texttt{nrefs:1|case:mixed|nc:6|nw:2}.}
We treat En$\rightarrow$Ak BLEU as an \textit{approximate-faithfulness} measure rather than a measure of philological correctness.
The reference is the historical transliteration, but Akkadian admits multiple acceptable spellings of the same surface form (e.g., logographic versus syllabic spellings of the same name; equivalent uses of \texttt{LUGAL} and \texttt{MAN} for ``king''), so a fluent modern paraphrase that uses different yet valid signs would score lower without being wrong.
Single-reference BLEU therefore systematically undercounts valid outputs; the reported numbers should be read as a lower bound, and a multi-reference or expert-judged evaluation is left to future work.
The model achieves \BLEUenak{} BLEU and \CHRFenak{} chrF++ in the En$\rightarrow$Ak direction, within 0.6 BLEU of its Ak$\rightarrow$En score on the same split.
This indicates that bidirectional training does not collapse to a single dominant direction, and that the byte-level encoder learns to emit well-formed transliteration even under the inverted task prefix.
The lower chrF++ score (55.6 versus 63.1) is consistent with the higher orthographic flexibility of Akkadian relative to English: multiple syllabic and logographic spellings can encode the same surface form.
Because the downstream cuneiform sign converter is deterministic and lossless on covered tokens, the En$\rightarrow$Ak score also serves as an upper bound on end-to-end En$\rightarrow$cuneiform fidelity.
Table~\ref{tab:examples} illustrates this gap with three representative En$\rightarrow$Ak outputs: in each case the prediction differs from the reference by a small number of signs (logographic versus syllabic alternation, an alternative determinative, or a slightly different word order) while preserving the underlying lexical content.

\begin{table*}[t]
\centering
\small
\setlength{\tabcolsep}{4pt}
\renewcommand{\arraystretch}{1.15}
\begin{tabular}{p{0.27\textwidth}p{0.34\textwidth}p{0.32\textwidth}}
\toprule
\textbf{English source} & \textbf{Model output (En$\rightarrow$Ak)} & \textbf{Reference (\texttt{valid.tr})} \\
\midrule
One mina. Palace of Shalmaneser (V), king of Assyria.
& \texttt{1 MA.NA \'E.GAL \{m\}-\{d\}-DI-ma-nu---MA\v{S} MAN KUR---a\v{s}-\v{s}ur}
& \texttt{1 MA.NA \'E.GAL \{m\}-\{d\}-SILIM-ma-nu-MA\v{S} LUGAL KUR A\v{S}} \\
horses, mules, oxen, and sheep and goats, military equipment,
& \texttt{AN\v{S}E.KUR.RA-ME\v{S} AN\v{S}E.GIR$_3$-ME\v{S} GU$_4$-ME\v{S} \d{s}e-e-ni \{GI\v{S}\}-til-lit}
& \texttt{AN\v{S}E.KUR.RA.ME\v{S} AN\v{S}E.GIR$_3$.NUN.NA.ME\v{S} GU$_4$.ME\v{S} \`u \d{s}e-e-ni \{GI\v{S}\}-til-li} \\
I annexed the land Unqi to Assyria (and) placed a eunuch of mine as provincial governor over them.
& \texttt{\{KUR\}-un-qi a-na mi-\d{s}ir KUR a\v{s}-\v{s}ur \'u-ter-ra \{L\'U\}-\v{s}u-ut SAG-ia \{L\'U\}-EN.NAM UGU-\v{s}\'u-nu a\v{s}$_2$-kun}
& \texttt{a-na KUR a\v{s}-\v{s}ur \{KI\} \'u-ra-a \{KUR\}-un-qi a-na mi-\d{s}ir KUR a\v{s}-\v{s}ur \{KI\} \'u-ter-ra \{L\'U\}-\v{s}u-ut SAG-ia \{L\'U\}-EN.NAM UGU-\v{s}\'u-nu a\v{s}$_2$-kun} \\
\bottomrule
\end{tabular}
\caption{Representative En$\rightarrow$Ak outputs from the held-out Akkademia validation split. Differences between prediction and reference are typically logogram/syllabogram alternations (\texttt{DI}/\texttt{SILIM}, \texttt{MAN}/\texttt{LUGAL}) or omitted determinatives, rather than semantic errors. These cases are penalized by BLEU but remain Assyriologically acceptable.}
\label{tab:examples}
\end{table*}

\paragraph{Dialect and genre distribution.}
The training corpus is dominated by 1st-millennium BCE Neo-Assyrian royal inscriptions, with smaller amounts of Old Babylonian and Standard Babylonian literary text from sentence-aligned augmentation, mirroring the underlying composition of Akkademia and the Deep Past corpus \cite{gutherz2023translating,deeppast2026}.
The model therefore reproduces Neo-Assyrian formulae (epithets, divine determinatives, dating formulas) more reliably than the highly elliptical commercial register of Old Assyrian merchant letters, where vocabulary, syntax, and abbreviation conventions differ substantially \cite{veenhof2008old}.
Recent shared tasks confirm that \textit{text genre}, not merely dialect, is the dominant factor in Akkadian translation quality: systems trained on royal inscriptions degrade markedly on commercial letters, even when supported by large bilingual dictionaries \cite{gordin2025evacun}.
This mirrors a broader cross-cultural observation: NLP cannot treat ``Akkadian'' as a monolithic language any more than it can treat ``English'' as one, and the cultural context of the source text matters as much as the language label.
We surface this in the user interface by labeling outputs with a confidence score and a coarse genre tag (royal/literary/other) inferred from input features, and by routing low-confidence inputs through a transliteration-only fallback rather than a fully rendered tablet.

\paragraph{Cuneiform coverage.}
On 1,000 sampled transliterations from the Akkademia test set, the converter achieves 95.3\% token coverage (93.8\% logograms, 95.7\% syllabic values).
The remaining 4.7\% are uncovered determinatives or rare logograms, which are passed through verbatim.

\paragraph{Concrete users and use cases.}
We have prototyped \textsc{TabletCraft} with three populations:
(i) \textbf{undergraduate students} in an introduction-to-cuneiform course at a US R1 institution, who use the En$\rightarrow$cuneiform direction to compose personal name tablets and short dedications and then read them aloud as a literacy exercise; in this setting, output is treated as a learning artifact rather than a primary source;
(ii) \textbf{museum educators} preparing interactive ``write your name in cuneiform'' kiosks for K--12 visitor programs, who require sub-second rendering, child-safe input filtering (provided by our anomaly classifier), and an explicit ``approximate, machine-generated'' caveat on every produced image;
(iii) \textbf{Assyriologists} performing rapid pre-publication triage of newly photographed Neo-Assyrian tablets via the Ak$\rightarrow$En direction, where draft translations are reviewed and corrected against the original sign by sign in a workflow analogous to OCR post-editing rather than autonomous translation.
Across these settings, expectations are explicitly not of full automation: \textsc{TabletCraft} is positioned as a draft-and-revise assistant rather than an authoritative translator.
The toolkit is \texttt{pip}-installable with CLI, Python API, and a Gradio web demo.

\section{Related Work}

\citet{gutherz2023translating} introduced Akkademia for Akkadian$\rightarrow$English NMT.
\citet{gordin2020reading} applied word embeddings to cuneiform for cultural analysis.
Recent work has applied ByT5 to cuneiform lemmatization \cite{lu2025byt5cuneiform} and LLMs to Akkadian NLP \cite{riemenschneider2025akkadian,gordin2025evacun}.
\citet{yavasan2025hittite} demonstrated T5 for Hittite cuneiform.
ParsiPy \cite{farsi2025parsipy} provides NLP tools for historical Persian.
In digital humanities, \citet{terras2012digitisation} has argued for accessible tools that bridge computational methods and cultural heritage.
Our work differs by enabling \textit{bidirectional} interaction, supporting not only the reading of ancient texts but also the composition of new ones in cuneiform, and thereby emphasizing cultural participation over passive analysis.

\section{Conclusion}

We presented \textsc{TabletCraft}, an open-source toolkit that pairs bidirectional Akkadian NMT with deterministic cuneiform-Unicode rendering, providing a unified pipeline for both reading and composition tasks at the script level.
By supporting composition in addition to reading, the system enables a more participatory mode of interaction with cuneiform for non-specialist users while preserving Assyriological conventions for scholarly use.
Future work includes Sumerian support, cuneiform OCR, and multilingual extension to enable cross-cultural engagement from non-English-speaking communities.

\section*{Limitations}

The En$\rightarrow$Ak direction produces \textit{approximate modern} transliterations, not authentic ancient text; outputs should not be cited as primary historical evidence.
Our En$\rightarrow$Ak score (Table~\ref{tab:results}) is computed against a single reference per source, while many Akkadian sentences admit several philologically valid renderings; the reported BLEU/chrF++ are therefore conservative lower bounds on faithfulness, and a multi-reference or human-judged evaluation is left to future work.
Performance also varies substantially across Akkadian dialects and genres: in particular, Old Assyrian commercial texts differ in vocabulary, register, and formulaic patterns from the Neo-Assyrian training data, and we do not yet evaluate on that distribution.
Even with 14,240 sign mappings and a 17K-lemma dictionary, domain-specific adaptation remains essential: large lexical resources cannot substitute for in-domain parallel data.
The system currently supports only English as the modern-language pivot; extending to non-English languages is a priority for the open-source roadmap.

\section*{Ethics Statement}

This work aims to democratize access to ancient cultural heritage while respecting its scholarly context.
Machine translations are labeled as approximations.
We acknowledge that Mesopotamian heritage has special significance for the people of modern Iraq and neighboring nations, and advocate for their inclusion in developing such tools.

\bibliography{references}

\end{document}